%% file: acl_latex.tex
\pdfoutput=1

\documentclass[11pt]{article}

\usepackage[]{acl}

\usepackage{times}
\usepackage{latexsym}
\usepackage{amssymb}
\usepackage{amsmath}
\usepackage{multirow}
\usepackage{balance}

\usepackage[T1]{fontenc}

\usepackage[utf8]{inputenc}

\usepackage{microtype}

\title{{ Trading Syntax Trees for Wordpieces:}\\ Target-oriented Opinion Words Extraction with Wordpieces and Aspect Enhancement }

  \author{Samuel Mensah \\
  Computer Science Department \\
  University of Sheffield, UK \\
  \texttt{\small s.mensah@sheffield.ac.uk} \\\And
  Kai Sun \\
  BDBC and SKLSDE \\
  Beihang University, China \\
  \texttt{\small sunkai@buaa.edu.cn} \\\And
  Nikolaos Aletras \\
  Computer Science Department \\
  University of Sheffield, UK \\
  \texttt{\small n.aletras@sheffield.ac.uk} \\}

\begin{document}
\maketitle

\begin{abstract}
State-of-the-art target-oriented opinion word extraction (TOWE) models typically use BERT-based text encoders that operate on the word level, along with graph convolutional networks (GCNs) that incorporate syntactic information extracted from syntax trees. These methods achieve limited gains with GCNs and have difficulty using BERT wordpieces. Meanwhile, BERT wordpieces are known to be effective at representing rare words or words with insufficient context information. To address this issue, this work trades syntax trees for BERT wordpieces by entirely removing the GCN component from the methods' architectures. To enhance TOWE performance, we tackle the issue of aspect representation loss during encoding. Instead of solely utilizing a sentence as the input, we use a sentence-aspect pair. Our relatively simple approach achieves state-of-the-art results on benchmark datasets and should serve as a strong baseline for further research.


\end{abstract}

\section{Introduction}

Target-oriented opinion word extraction (TOWE; \citet{fan2019target}) is a subtask in aspect-based sentiment analysis (ABSA; \citet{jphpiasm2014semeval}), which aims to identify words that express an opinion about a specific target (or aspect) in a sentence. For instance, in the sentence { ``Such an {\em awesome} {\bf surfboard}.''}, a TOWE model should identify { \em``awesome''} as the opinion word for the given aspect {\bf surfboard}. TOWE provides explicit aspect-opinion pairs which can be used to improve results in downstream tasks such as opinion summarization~\cite{kim2011comprehensive} and information extraction~\cite{jphpiasm2014semeval,tang2016aspect,sun2023self}.

%



Currently, many TOWE methods~\cite{DBLP:conf/emnlp/VeysehNDDN20,SDRN2020,ARGCN2021,TSMSA2021,mensah2021empirical} use pretrained BERT~\cite{devlin2018bert}, to encode the input sentence. BERT has the ability to effectively capture context, which can improve TOWE performance. However, many of these methods are rather complex, as they often incorporate syntax tree information using a graph convolutional network (GCN)~\cite{kipf2016semi}. For instance, \citet{DBLP:conf/emnlp/VeysehNDDN20} uses an ordered-neuron LSTM \cite{shen2018ordered} encoder with a  GCN while \citet{ARGCN2021} applies an attention-based relational GCN on the syntax tree. \citet{mensah2021empirical} applies a BiLSTM \cite{hochreiter1997long} on BERT embeddings and incoporate syntax information via a  GCN.

\renewcommand*{\arraystretch}{1.3}
\begin{table}[t]
\small
    \centering

    \begin{tabular}{@{}ll@{}}
    \hline
       {1. \bf Sentence}:& Such an {\em awesome} {\bf surfboard}    \\ 
       {\bf \,\,\,\, Wordpieces:}& `such', `an', {\em `awesome'}, {\bf `surf'}, \\ & {\bf `\#\#board'}  \\ \hline
       {2. \bf Sentence}:& A {\em great} {\bf snowboard} which holds edges \\ 
       & well  when riding on snow.    \\ 
       {\bf \,\,\,\, Wordpieces:}&  `A', {\em `great'}, {\bf `snow'}, {\bf `\#\#board'},  
        `which', \\ 
        & `holds', `edges', `well', `when', `riding', \\ 
        & `on', `snow'.\\ \hline  
          
    \end{tabular}
    \caption{Sentences demonstrating contextual understanding through shared wordpieces. The table shows each sentence and its corresponding BERT wordpiece sequence. Aspect words are bold-typed and opinion words are italicized. The shared wordpiece '\#\#board' helps in decoding the meaning of ``surfboard''.   }
    \label{tab:example}
\end{table}

While incorporating syntax information through GCNs has been shown to provide some performance gains in TOWE, these are usually limited~\cite{mensah2021empirical}. Moreover, modeling subword tokens with a GCN can be challenging because the syntax tree consists of whole words rather than subword tokens like wordpieces~\cite{schuster2012japanese,devlin2018bert}. Models based on subword tokens strike a good balance between character- and word-based encoders. They are able to effectively learn representations of rare words or words with insufficient context information. Consider the example in Table~\ref{tab:example}. The context information for ``surfboard'' is limited, making it difficult to understand its meaning without additional context. However, both aspects share the wordpiece ``\#\#board'', which allows the meaning of ``surfboard'' to be partially understood by using information from the context of ``snowboard''. In this case, ``riding'' is related to both aspects through the shared wordpiece, enabling the representation of ``surfboard'' to be improved. 


In this paper, we propose a substantial simplification for syntax-aware TOWE models \cite{DBLP:conf/emnlp/VeysehNDDN20,ARGCN2021,mensah2021empirical} by replacing the syntax tree with subword information while maintaining good prediction performance. This is accomplished by removing the GCN from these architectures and using BERT wordpieces instead. Additionally, we address the issue of aspect representation degradation during encoding. This degradation negatively affects TOWE performance by reducing the availability of semantic information about the aspect for determining the opinion words to extract. To solve this problem, we propose using a sentence-aspect pair as input rather than just a sentence, similar to the approach used by \citet{tian2021aspect} for aspect-based sentiment classification. Through extensive experimentation, we found that our simple approach achieves state-of-the-art (SOTA) results by outperforming the method proposed by \citet{mensah2021empirical} without the need of a GCN component.

\section{Task Formalization}

The TOWE task aims to identify an opinion word in a sentence $S=\{w_1,\ldots,w_{n_s}\}$ with respect to an aspect $w_a\in S$. The sentence is typically tokenized into a sequence of tokens at different levels of granularity (e.g. subwords or whole words), $T=\{t_1,\ldots,t_{n_t}\}$, with $t_a\in T$ denoting a sub-sequence of the aspect $w_a$ and $n_s\leq n_t$. The goal is to assign one of three tags (I, O, or B) to each token using the IOB format~\cite{DBLP:conf/acl-vlc/RamshawM95}, which indicates whether the word is at the Inside, Outside or Beginning of the opinion word relative to the aspect.



\section{Syntax-aware Approaches to TOWE}\label{sec:syntax-models}



Typically, syntax-aware approaches to TOWE~\cite{DBLP:conf/emnlp/VeysehNDDN20,ARGCN2021,mensah2021empirical} employ a text encoder that utilizes pre-trained BERT \cite{devlin2018bert} and position embeddings \cite{zeng2014relation} (or category embeddings \cite{ARGCN2021}) to learn whole word representations that are aware of the aspect's location in text. These approaches also include a GCN that operates on a syntax tree in order to incorporate syntactic information into the model.


%


\paragraph{Ordered-Neuron LSTM GCN (ONG):} \citet{DBLP:conf/emnlp/VeysehNDDN20} combine an ordered neuron LSTM (ON-LSTM; \citet{shen2018ordered}) and a GCN for TOWE. The ON-LSTM layer is an LSTM variant that considers the order of elements in the input sequence (including BERT and position embeddings) when modeling dependencies between them. The GCN encodes syntactic structural information into the representations obtained by the ON-LSTM layer.

\paragraph{BERT+BiLSTM+GCN:} \citet{mensah2021empirical} replaces the ON-LSTM of the ONG model with a BiLSTM to better capture short-term dependencies between aspect and opinion words.


\paragraph{Attention-based Relational GCN (ARGCN):} \citet{ARGCN2021} combine contextualized embedding obtained using BERT with a category embedding (i.e., IOB tag embedding) to incorporate aspect information. They subsequently use a relational GCN \cite{schlichtkrull2018modeling} and BiLSTM to respectively incorporate syntactic and sequential information for TOWE classification.

\section{Trading Syntax Trees for Wordpieces}\label{sec:proposed-methodology}

\citet{mensah2021empirical} have recently demonstrated that the use of a GCN to incorporate syntax tree information has little impact in TOWE model performance. Meanwhile, the GCN presents challenges when using subword tokens, as previously mentioned. Therefore, we propose a simplified version of the TOWE model that omits the GCN component from syntax-aware approaches and instead uses subword tokens as the input to the BERT component. In this work, we use BERT's Wordpieces~\cite{devlin2018bert} as the subword representation because they are highly informative, having been derived from the BERT pretraining process. However, methods such as Byte-Pair Encoding (BPE)~\cite{sennrich2015neural} can also be used, as we will see later in the experiments.




\renewcommand*{\arraystretch}{1.3}
\begin{table*}[t]
\centering
    \small
    \centering

    \begin{tabular}{l|c|c|c|c|c||c} \hline
      \bf Models    &\bf Granularity&\bf Lap14 &\bf Res14 &\bf Res15 &\bf Res16 & \bf Avg \\\hline
     ONG   & word  & 75.77 & 82.33 & 78.81 & 86.01 & 80.73   \\
                             ONG w/o GCN   &word& 74.17 &84.10 &78.33 &84.87 & 80.37   \\
                               ONG(S) w/o GCN  &wordpiece& 79.79 &86.63 &80.72 &88.30 & 83.86  \\
                               ONG(S,A) w/o GCN  &wordpiece& 81.70 &\bf 88.70 &\bf 82.55 &91.18 & 86.03\\ \hline 
     ARGCN    &word & 76.36 &85.42 &78.24 &86.69 & 81.68   \\
                                ARGCN w/o R-GCN   &word & 76.38 &84.36 &78.41 &84.61 & 80.94   \\
                               ARGCN(S) w/o R-GCN  &wordpiece& 80.08 &85.92 &81.36 &89.72  & 84.27 \\
                             ARGCN(S,A) w/o R-GCN &wordpiece& 81.37 &88.18 & 82.49 &90.82 & 85.72\\ \hline 
      BERT+BiLSTM+GCN   &word & 78.82 &85.74 &80.54 & 87.35 & 83.11 \\
                               BERT+BiLSTM   &word & 78.25 &  85.60 &  80.41 & 86.94 & 82.80 \\
                               BERT+BiLSTM(S)  &wordpiece& 80.45	&86.27		&80.89	&89.80 & 84.35    \\
                               BERT+BiLSTM(S,A) &wordpiece& \bf 82.59	&88.60 &82.37 &\bf91.25 & \bf 86.20 \\ \hline 

    \end{tabular}
    \caption{F1 performance of syntax-aware methods and their variants. "Avg" refers to the average F1 score calculated across all of the datasets. ``Granularity'' highlights the level of granularity at which input tokens are represented.}
    \label{tab:my_label}
\end{table*}

\subsection{Formatting BERT Input}
Given sentence $S$, the BERT wordpiece tokenizer segments $S$ into a sequence of wordpieces $T=\{t_1,t_2,\ldots,t_{n_t}\}$. The BERT input for $S$ is then formatted as follows:
\begin{equation}
     T^{(S)}=\{ \textsc{[CLS]} , T , \textsc{[SEP]} \}
\end{equation}
\noindent where \textsc{[CLS]} and \textsc{[SEP]} are special tokens that mark the boundaries of the sentence. 

While this format may be adequate for some NLP tasks, it can be problematic for learning good aspect representations in aspect-based sentiment classifica-
tion ~\cite{tian2021aspect}. To mitigate this issue, we adopt the approach of \citet{tian2021aspect} and reformat the BERT input by using a sentence-aspect pair $T^{(S,A)}$, which combines $T^{(S)}$ and $t_a$ (i.e. the aspect subsequence) along with special tokens.
\begin{equation}
     T^{(S,A)}=\{ \textsc{[CLS]} , T , \textsc{[SEP]}, t_a, \textsc{[SEP]} \}
\end{equation}

\subsection{Classification and Optimization}

The input $T^{(S,A)}$ consists of two parts: $T^{(S)}$ and $t_a$. Since $t_a$ only serves to enhance the aspect representation in $T^{(S)}$, sequence labeling is done on $T^{(S)}$ only. During sequence labeling, we follow the common approach of predicting based on the first wordpiece representation of a word. For instance, given the word ``surfboard'' that consists of the wordpieces ``surf'' and ``\#\#board'' which both are learned  during encoding, only the representation of ``surf'' is fed to a softmax classifier to predict the tag for the whole word. The cross-entropy function is minimized for each word in the training set.

\section{Experiments and Results}\label{sec:exp-results}

We experiment with the following baselines: ARGCN, BERT+BiLSTM+GCN and ONG. We use the suffixes (S) or (S,A) to indicate whether the modified versions of these methods uses a wordpiece sentence or wordpiece sentence-aspect pair as input, respectively. We used the publicly available code and optimal hyperparameter settings from the authors of ARGCN\footnote{\url{https://github.com/samensah/encoders_towe_emnlp2021}} and BERT+BiLSTM+GCN.\footnote{\url{https://github.com/wcwowwwww/towe-eacl}} We have implemented ONG model variants ourselves using the suggested hyperparameter configurations from the authors.\footnote{\url{https://github.com/samensah/Towe-TradeSyntax4WP}} Following previous work \cite{fan2019target}, we use the same experimental setup and evaluate on the Laptop dataset (Lap14) and the Restaurant datasets (Res14, Res15, Res16) \cite{pontiki2014semeval,pontiki2015semeval,pontiki2016semeval}. The result reported for each dataset is the average over Micro F1 scores obtained from five different runs. Each run uses a different random seed to ensure the stability of our results. 

\subsection{F1 Performance Comparison}
The results, shown in Table \ref{tab:my_label}, indicate that removing the GCN component from syntax-aware approaches does not substantially impact their performance, with average decreases in performance of 0.36, 0.74, and 0.31, respectively. However, we observed a large improvement in model performance when using wordpieces, as indicated by the models with the (S) suffix. It is possible that BERT captures enough syntax information already \cite{clark-etal-2019-bert} and, therefore, using GCNs to exploit syntax trees does not substantially improve performance on the task. This suggests that it may be beneficial to prioritize wordpieces over syntax trees to allow BERT to fully utilize rare and out-of-vocabulary words. We also discovered that using a sentence-aspect pair as input resulted in better performance than using only the sentence for the models, as indicated by the results of models with the (S,A) suffix.  We believe that this may be due to the aspect information being lost or degraded during the encoding process for models with the (S) suffix. Among the methods, BERT+BiLSTM(S,A) had the highest average F1 score of 86.2.

\subsection{Influence of Aspect Representation}

\begin{table}[t]
\centering
\scriptsize
\begin{tabular}{l|c|c|c|c||c}\hline
{\bf Model}  &\bf  Lap14 &\bf   Res14 &\bf  Res15 &\bf  Res16 & \bf Avg \\ \hline %


BERT-BiLSTM(S) &80.45	&86.27		&80.89	&89.80 & 84.35 \\
-Mask Aspect  &80.01  &86.11 	&80.42  &88.59 & 83.78 \\

\hline

\end{tabular}
\caption{F1 performance of BERT-BiLSTM(S) with and without aspect masking.}
\label{table:aspect-oriented}
\end{table}




To determine if the aspect representation is degraded during encoding, we evaluate BERT+BiLSTM(S) with and without aspect masking. The results, shown in Table \ref{table:aspect-oriented}, show that masking the aspect representation had only a minimal impact on performance, with a decrease in performance of 0.44 (Lap14), 0.16 (Res14), 0.47 (Res15), and 1.2 (Res16). These findings suggest that the aspect information has limited contribution and requires enhancement to improve performance, as demonstrated by the improved results of BERT+BiLSTM(S,A).



\subsection{Qualitative Analysis}

\renewcommand*{\arraystretch}{1.3}
\begin{table*}[t]
    \small
    \begin{tabular}{p{5.5cm}|p{2.5cm}|p{2.5cm}|p{2.5cm}}
    \hline
         \bf Sentence  & \bf BERT+BiLSTM & \bf BERT+BiLSTM(S) & \bf BERT+BiLSTM(S,A) \\
        \hline
         The \textbf{OS} is \textit{fast} and \textit{fluid}, everything is organized and it's just \textit{beautiful}.  & \it fast, fluid & \it fast, fluid& \it fast,  fluid, beautiful \\\hline

         Certainly \textit{not the best} \textbf{sushi} in new york, however, it is always \textit{fresh}, and the place is very clean, sterile. & \it fresh & \it not the best& \it not the best, fresh \\\hline
         
         Although somewhat load, the \textbf{noise} was \textit{minimally intrusive} & \it loud, intrusive & \it loud, minimally intrusive & \it loud, minimally intrusive.\\\hline
         
         \end{tabular}
         \caption{Case Study: Evaluating the model performance on different case examples. Aspect words are bold-typed and opinion words are italicized.}
         \label{case-study}
\end{table*}


We examined the performance of BERT+BiLSTM, BERT+BiLSTM(S), and BERT+BiLSTM(S,A) on three case examples, as shown in Table~\ref{case-study}. The results show that the BERT+BiLSTM and BERT+BiLSTM(S) models struggled to identify opinion words that were farther away from the aspect, particularly in the first and second cases where the opinion words ``beautiful'' and ``fresh'' were missed. Upon further investigation, we discovered that these opinion words were closer to the aspect's co-referential term ``it''. The model struggled to determine what ``it'' referred to due to degradation of the aspect representation, leading to the missed identification of the opinion words. However, BERT+BiLSTM(S,A) was able to recover these opinion words due to its ability to enhance the aspect representation. In the third case example, the use of wordpieces was beneficial as the opinion word ``minimally’’ was not present in the training set, but its wordpiece ``\#\#ly,’’ was associated with 15 opinion words in the training set. BERT+BiLSTM(S) and BERT+BiLSTM(S,A) were able to identify the opinion word ``minimally’’ in the test set by leveraging the context of ``\#\#ly,’’.


\section{Impact of BPE Subword Representations}\label{sec:appendix}

\begin{table}[t]
\centering
\scriptsize
\begin{tabular}{l|c|c|c|c||c}\hline
{\bf Model}  &\bf  Lap14 &\bf   Res14 &\bf  Res15 &\bf  Res16 & \bf Avg \\ \hline %

{\tiny RoBERTa-BiLSTM(S,A)} &82.77 &88.27 &83.84 &91.06 & 86.49\\
{\tiny RoBERTa-BiLSTM(S)} &81.10 &86.95 &82.21 &88.70 & 84.74\\
{\tiny RoBERTa-BiLSTM} &75.87 &81.38 &75.94 &84.70 & 79.47\\
{\tiny RoBERTa-BiLSTM+GCN} &77.57 &82.09 &77.85 &85.37 & 80.72 \\
\hline

\end{tabular}
\caption{F1 Performance of RoBERTa models to investigate the use of BPE subword representations.}
\label{table:roberta}
\end{table}

We previously examined the use of wordpiece representations derived from pretrained BERT for TOWE models. In this section, we look into using Byte Pair Encoding (BPE)~\cite{sennrich2015neural} as an alternative method for subword representation, which is inspired by data compression techniques \cite{gage1994new}. It is worth noting that BPE representations are generally not obtained from pretrained BERT. However, since RoBERTa is pretrained using BPE, and RoBERTa is a variant of BERT, we can still explore the impact of using BPE representations in TOWE models. To do this, we replace the BERT component in our best model, BERT+BiLSTM(S,A), with RoBERTa, developing the model RoBERTa+BiLSTM(S,A). The results of RoBERTa+BiLSTM(S,A) and its variations are shown in Table~\ref{table:roberta}.

Note, while RoBERTa+BiLSTM(S,A) and RoBERTa+BiLSTM(S) use BPE subword token representations as input, RoBERTa+BiLSTM and RoBERTa+BiLSTM+GCN operate on the word-level. Our findings support the notion that GCNs have a limited impact on performance, as demonstrated by a relatively small decrease in average F1 score when comparing RoBERTa+BiLSTM+GCN to RoBERTa+BiLSTM. On the other hand, using BPE representations instead of GCN resulted in a substantial improvement in model performance of +5.27 when comparing RoBERTa+BiLSTM and RoBERTa+BiLSTM(S). The results indicate that syntax trees via GCNs may not be necessary and can be replaced by subword representations such as BPE for better performance in TOWE. Additionally, the performance of RoBERTa+BiLSTM(S) can be further improved by using BPE-based sentence-aspect pairs, as seen by the +1.75 performance gain in RoBERTa+BiLSTM(S,A).

\subsection{State-of-the-art Models}

\input{all_methods}

Finally, we compare the performance of BERT+BiLSTM(S,A) with recent methods, including IOG~\cite{fan2019target}, LOTN~\cite{DBLP:conf/aaai/WuZDHC20}, SDRN+BERT~\cite{SDRN2020}, BERT+BiLSTM+GCN \cite{mensah2021empirical}, QD-OWSE~\cite{gao2021question}, TSMSA~\cite{TSMSA2021}. The results of this comparison are shown in Table~\ref{table:SOTA}. Among these methods, the recent proposed methods QD-OWSE and TSMSA, which both use BERT as a basis for their approach, achieved competitive results with ours. QD-OWSE uses a generated question-answer pair as BERT input, while TSMSA uses multi-head attention to identify opinion words. These methods go on to demonstrate that BERT can capture sufficient syntax information for this task, even without the use of syntax trees. However, BERT+BiLSTM(S,A) achieved the best results, with F1 scores 82.59 (Lap14), 88.6 (Res14), 82.37 (Res15) and 91.25 (Res16),  setting a new SOTA for the task.

\section{Conclusion}\label{sec:conclusion}

We demonstrated that replacing GCNs with BERT wordpieces while enhancing the aspect representation achieves SOTA results in syntax-aware TOWE approaches. 
The aspect enhancement method serves as a ``prompt'' for the model. 
We intend to explore prompt-based learning \cite{brown2020language} to further improve the aspect representation.


\section{Limitations}

Currently, our approach does not effectively leverage syntax tree information via GCNs, a commonly used method for incorporating syntax trees in this task. Further research is required to determine the most effective way to integrate syntax tree information into TOWE models.


\section*{Acknowledgements}
This work was supported by the Leverhulme Trust under Grant Number: RPG\#2020\#148.

\balance

\bibliography{anthology,custom}
\bibliographystyle{acl_natbib}

\clearpage
\newpage

\appendix

\end{document}

%% file: all_methods.tex
\begin{table}[htbp]
\centering
\scriptsize
\renewcommand*{\arraystretch}{1.3}
\begin{tabular}{l|c|c|c|c||c}\hline
\bf Model &\bf Lap14 &\bf Res14 &\bf Res15 &\bf Res16 &\bf Avg\\   \hline
IOG  &  71.35  &80.02 &73.25 &81.69 & 76.58\\ %
LOTN &   72.02  &82.21 &73.29 & 83.62 & 77.79 \\ %

SDRN+BERT*  & 73.69  & 83.10 & 76.38 & 85.40 & 79.64 \\
ONG &  75.77   &82.33 &78.81 & 86.01 &  80.73 \\ 
ARGCN & 76.36 &85.42 &78.24 &86.69 & 81.68\\
BERT+BiLSTM+GCN &78.82 &85.74 &80.54 &87.35 & 83.11 \\
QD-OWSE &80.35 &87.23 &80.71 &88.14 & 84.11\\
TSMSA &82.18 &86.37 &81.64 &89.20 & 84.85 \\


 \hline
BERT-BiLSTM (S,A) &\bf 82.59	&\bf88.60 &\bf82.37 &\bf91.25 &\bf  86.20\\ \hline
\end{tabular}
\caption{Performance of TOWE methods. Results for the method marked ``*'' are from \cite{ARGCN2021}. }
\label{table:SOTA}
\end{table}

%% file: acl_latex.bbl
\begin{thebibliography}{28}
\expandafter\ifx\csname natexlab\endcsname\relax\def\natexlab#1{#1}\fi

\bibitem[{Brown et~al.(2020)Brown, Mann, Ryder, Subbiah, Kaplan, Dhariwal,
  Neelakantan, Shyam, Sastry, Askell et~al.}]{brown2020language}
Tom Brown, Benjamin Mann, Nick Ryder, Melanie Subbiah, Jared~D Kaplan, Prafulla
  Dhariwal, Arvind Neelakantan, Pranav Shyam, Girish Sastry, Amanda Askell,
  et~al. 2020.
\newblock Language models are few-shot learners.
\newblock \emph{Advances in neural information processing systems},
  33:1877--1901.

\bibitem[{Chen et~al.(2020)Chen, Liu, Wang, Zhang, and Chi}]{SDRN2020}
Shaowei Chen, Jie Liu, Yu~Wang, Wenzheng Zhang, and Ziming Chi. 2020.
\newblock Synchronous double-channel recurrent network for aspect-opinion pair
  extraction.
\newblock In \emph{Proceedings of the 58th Annual Meeting of the Association
  for Computational Linguistics}, pages 6515--6524.

\bibitem[{Clark et~al.(2019)Clark, Khandelwal, Levy, and
  Manning}]{clark-etal-2019-bert}
Kevin Clark, Urvashi Khandelwal, Omer Levy, and Christopher~D. Manning. 2019.
\newblock \href {https://doi.org/10.18653/v1/W19-4828} {What does {BERT} look
  at? an analysis of {BERT}{'}s attention}.
\newblock In \emph{Proceedings of the 2019 ACL Workshop BlackboxNLP: Analyzing
  and Interpreting Neural Networks for NLP}, pages 276--286, Florence, Italy.
  Association for Computational Linguistics.

\bibitem[{{Devlin} et~al.(2018){Devlin}, {Chang}, {Lee}, and
  {Toutanova}}]{devlin2018bert}
Jacob {Devlin}, Ming-Wei {Chang}, Kenton {Lee}, and Kristina~N. {Toutanova}.
  2018.
\newblock Bert: Pre-training of deep bidirectional transformers for language
  understanding.
\newblock In \emph{Proceedings of the 2019 Conference of the North American
  Chapter of the Association for Computational Linguistics: Human Language
  Technologies, Volume 1 (Long and Short Papers)}, pages 4171--4186.

\bibitem[{Fan et~al.(2019)Fan, Wu, Dai, Huang, and Chen}]{fan2019target}
Zhifang Fan, Zhen Wu, Xinyu Dai, Shujian Huang, and Jiajun Chen. 2019.
\newblock Target-oriented opinion words extraction with target-fused neural
  sequence labeling.
\newblock In \emph{Proceedings of the 2019 Conference of the North American
  Chapter of the Association for Computational Linguistics: Human Language
  Technologies, Volume 1 (Long and Short Papers)}, pages 2509--2518.

\bibitem[{Feng et~al.(2021)Feng, Rao, Tang, Wang, and Liu}]{TSMSA2021}
Yuhao Feng, Yanghui Rao, Yuyao Tang, Ninghua Wang, and He~Liu. 2021.
\newblock Target-specified sequence labeling with multi-head self-attention for
  target-oriented opinion words extraction.
\newblock In \emph{Proceedings of the 2021 Conference of the North American
  Chapter of the Association for Computational Linguistics: Human Language
  Technologies}, pages 1805--1815.

\bibitem[{Gage(1994)}]{gage1994new}
Philip Gage. 1994.
\newblock A new algorithm for data compression.
\newblock \emph{C Users Journal}, 12(2):23--38.

\bibitem[{Gao et~al.(2021)Gao, Wang, Liu, Wang, Zhang, and
  Liao}]{gao2021question}
Lei Gao, Yulong Wang, Tongcun Liu, Jingyu Wang, Lei Zhang, and Jianxin Liao.
  2021.
\newblock Question-driven span labeling model for aspect--opinion pair
  extraction.
\newblock In \emph{Proceedings of the AAAI Conference on Artificial
  Intelligence}, volume~35, pages 12875--12883.

\bibitem[{{Hochreiter} and {Schmidhuber}(1997)}]{hochreiter1997long}
Sepp {Hochreiter} and Jürgen {Schmidhuber}. 1997.
\newblock Long short-term memory.
\newblock \emph{Neural Computation}, 9(8):1735--1780.

\bibitem[{Jiang et~al.(2021)Jiang, Wang, and Aizawa}]{ARGCN2021}
Junfeng Jiang, An~Wang, and Akiko Aizawa. 2021.
\newblock Attention-based relational graph convolutional network for
  target-oriented opinion words extraction.
\newblock In \emph{Proceedings of the 16th Conference of the European Chapter
  of the Association for Computational Linguistics: Main Volume}, pages
  1986--1997.

\bibitem[{Kim et~al.(2011)Kim, Ganesan, Sondhi, and
  Zhai}]{kim2011comprehensive}
Hyun~Duk Kim, Kavita Ganesan, Parikshit Sondhi, and ChengXiang Zhai. 2011.
\newblock Comprehensive review of opinion summarization.

\bibitem[{Kipf and Welling(2017)}]{kipf2016semi}
Thomas~N. Kipf and Max Welling. 2017.
\newblock \href {https://openreview.net/forum?id=SJU4ayYgl} {Semi-supervised
  classification with graph convolutional networks}.
\newblock In \emph{5th International Conference on Learning Representations,
  {ICLR} 2017, Toulon, France, April 24-26, 2017, Conference Track
  Proceedings}. OpenReview.net.

\bibitem[{Mensah et~al.(2021)Mensah, Sun, and Aletras}]{mensah2021empirical}
Samuel Mensah, Kai Sun, and Nikolaos Aletras. 2021.
\newblock \href {https://aclanthology.org/2021.emnlp-main.722} {An empirical
  study on leveraging position embeddings for target-oriented opinion words
  extraction}.
\newblock In \emph{Proceedings of the 2021 Conference on Empirical Methods in
  Natural Language Processing}, pages 9174--9179, Online and Punta Cana,
  Dominican Republic. Association for Computational Linguistics.

\bibitem[{Pontiki et~al.(2016)Pontiki, Galanis, Papageorgiou, Androutsopoulos,
  Manandhar, Al-Smadi, Al-Ayyoub, Zhao, Qin, De~Clercq
  et~al.}]{pontiki2016semeval}
Maria Pontiki, Dimitrios Galanis, Haris Papageorgiou, Ion Androutsopoulos,
  Suresh Manandhar, Mohammad Al-Smadi, Mahmoud Al-Ayyoub, Yanyan Zhao, Bing
  Qin, Orph{\'e}e De~Clercq, et~al. 2016.
\newblock Semeval-2016 task 5: Aspect based sentiment analysis.
\newblock In \emph{International workshop on semantic evaluation}, pages
  19--30.

\bibitem[{Pontiki et~al.(2015)Pontiki, Galanis, Papageorgiou, Manandhar, and
  Androutsopoulos}]{pontiki2015semeval}
Maria Pontiki, Dimitrios Galanis, Harris Papageorgiou, Suresh Manandhar, and
  Ion Androutsopoulos. 2015.
\newblock Semeval-2015 task 12: Aspect based sentiment analysis.
\newblock In \emph{Proceedings of the 9th international workshop on semantic
  evaluation (SemEval 2015)}, pages 486--495.

\bibitem[{Pontiki et~al.(2014{\natexlab{a}})Pontiki, Galanis, Pavlopoulos,
  Papageorgiou, Androutsopoulos, and Manandhar}]{pontiki2014semeval}
Maria Pontiki, Dimitrios Galanis, John Pavlopoulos, Harris Papageorgiou, Ion
  Androutsopoulos, and Suresh Manandhar. 2014{\natexlab{a}}.
\newblock Semeval-2014 task 4: Aspect based sentiment analysis.
\newblock In \emph{Proceedings of the 8th International Workshop on Semantic
  Evaluation (SemEval 2014)}, page 27–35.

\bibitem[{Pontiki et~al.(2014{\natexlab{b}})Pontiki, Galanis, Pavlopoulos,
  Papageorgiou, Androutsopoulos, and Manandhar}]{jphpiasm2014semeval}
Maria Pontiki, Dimitris Galanis, John Pavlopoulos, Harris Papageorgiou, Ion
  Androutsopoulos, and Suresh Manandhar. 2014{\natexlab{b}}.
\newblock \href {https://doi.org/10.3115/v1/s14-2004} {Semeval-2014 task 4:
  Aspect based sentiment analysis}.
\newblock In \emph{Proceedings of the 8th International Workshop on Semantic
  Evaluation, SemEval@COLING 2014, Dublin, Ireland, August 23-24, 2014}, pages
  27--35. The Association for Computer Linguistics.

\bibitem[{Ramshaw and Marcus(1995)}]{DBLP:conf/acl-vlc/RamshawM95}
Lance~A. Ramshaw and Mitch Marcus. 1995.
\newblock \href {https://aclanthology.org/W95-0107/} {Text chunking using
  transformation-based learning}.
\newblock In \emph{Third Workshop on Very Large Corpora, VLC@ACL 1995,
  Cambridge, Massachusetts, USA, June 30, 1995}.

\bibitem[{Schlichtkrull et~al.(2018)Schlichtkrull, Kipf, Bloem, van den Berg,
  Titov, and Welling}]{schlichtkrull2018modeling}
Michael Schlichtkrull, Thomas~N. Kipf, Peter Bloem, Rianne van den Berg, Ivan
  Titov, and Max Welling. 2018.
\newblock Modeling relational data with graph convolutional networks.
\newblock In \emph{The Semantic Web}, pages 593--607, Cham. Springer
  International Publishing.

\bibitem[{Schuster and Nakajima(2012)}]{schuster2012japanese}
Mike Schuster and Kaisuke Nakajima. 2012.
\newblock Japanese and korean voice search.
\newblock In \emph{2012 IEEE International Conference on Acoustics, Speech and
  Signal Processing (ICASSP)}, pages 5149--5152. IEEE.

\bibitem[{Sennrich et~al.(2016)Sennrich, Haddow, and
  Birch}]{sennrich2015neural}
Rico Sennrich, Barry Haddow, and Alexandra Birch. 2016.
\newblock \href {https://doi.org/10.18653/v1/p16-1162} {Neural machine
  translation of rare words with subword units}.
\newblock In \emph{Proceedings of the 54th Annual Meeting of the Association
  for Computational Linguistics, {ACL} 2016, August 7-12, 2016, Berlin,
  Germany, Volume 1: Long Papers}. The Association for Computer Linguistics.

\bibitem[{{Shen} et~al.(2018){Shen}, {Tan}, {Sordoni}, and
  {Courville}}]{shen2018ordered}
Yikang {Shen}, Shawn {Tan}, Alessandro {Sordoni}, and Aaron~C. {Courville}.
  2018.
\newblock Ordered neurons: Integrating tree structures into recurrent neural
  networks.
\newblock In \emph{International Conference on Learning Representations}.

\bibitem[{Sun et~al.(2023)Sun, Zhang, Samuel, Nikolaos, Mao, and
  Liu}]{sun2023self}
Kai Sun, Richong Zhang, Mensah Samuel, Aletras Nikolaos, Yongyi Mao, and Xudong
  Liu. 2023.
\newblock Self-training through classifier disagreement for cross-domain
  opinion target extraction.
\newblock In \emph{Proceedings of the ACM Web Conference 2023}, pages
  1594--1603.

\bibitem[{{Tang} et~al.(2016){Tang}, {Qin}, and {Liu}}]{tang2016aspect}
Duyu {Tang}, Bing {Qin}, and Ting {Liu}. 2016.
\newblock Aspect level sentiment classification with deep memory network.
\newblock In \emph{Proceedings of the 2016 Conference on Empirical Methods in
  Natural Language Processing}, pages 214--224.

\bibitem[{Tian et~al.(2021)Tian, Chen, and Song}]{tian2021aspect}
Yuanhe Tian, Guimin Chen, and Yan Song. 2021.
\newblock Aspect-based sentiment analysis with type-aware graph convolutional
  networks and layer ensemble.
\newblock In \emph{Proceedings of the 2021 Conference of the North American
  Chapter of the Association for Computational Linguistics: Human Language
  Technologies}, pages 2910--2922.

\bibitem[{Veyseh et~al.(2020)Veyseh, Nouri, Dernoncourt, Dou, and
  Nguyen}]{DBLP:conf/emnlp/VeysehNDDN20}
Amir Pouran~Ben Veyseh, Nasim Nouri, Franck Dernoncourt, Dejing Dou, and
  Thien~Huu Nguyen. 2020.
\newblock \href {https://doi.org/10.18653/v1/2020.emnlp-main.719} {Introducing
  syntactic structures into target opinion word extraction with deep learning}.
\newblock In \emph{Proceedings of the 2020 Conference on Empirical Methods in
  Natural Language Processing, {EMNLP} 2020, Online, November 16-20, 2020},
  pages 8947--8956. Association for Computational Linguistics.

\bibitem[{Wu et~al.(2020)Wu, Zhao, Dai, Huang, and
  Chen}]{DBLP:conf/aaai/WuZDHC20}
Zhen Wu, Fei Zhao, Xin{-}Yu Dai, Shujian Huang, and Jiajun Chen. 2020.
\newblock \href {https://aaai.org/ojs/index.php/AAAI/article/view/6469} {Latent
  opinions transfer network for target-oriented opinion words extraction}.
\newblock In \emph{The Thirty-Fourth {AAAI} Conference on Artificial
  Intelligence, {AAAI} 2020, The Thirty-Second Innovative Applications of
  Artificial Intelligence Conference, {IAAI} 2020, The Tenth {AAAI} Symposium
  on Educational Advances in Artificial Intelligence, {EAAI} 2020, New York,
  NY, USA, February 7-12, 2020}, pages 9298--9305. {AAAI} Press.

\bibitem[{Zeng et~al.(2014)Zeng, Liu, Lai, Zhou, and Zhao}]{zeng2014relation}
Daojian Zeng, Kang Liu, Siwei Lai, Guangyou Zhou, and Jun Zhao. 2014.
\newblock Relation classification via convolutional deep neural network.
\newblock In \emph{Proceedings of COLING 2014, the 25th international
  conference on computational linguistics: technical papers}, pages 2335--2344.

\end{thebibliography}
